\documentclass[10pt,twocolumn,letterpaper]{article}

\usepackage{cvpr}
\usepackage{times}
\usepackage{epsfig}
\usepackage{graphicx}
\usepackage{amsmath}
\usepackage{amssymb}

\usepackage{enumitem}
\usepackage{booktabs} 
\usepackage{xcolor}

\usepackage[breaklinks=true,bookmarks=false]{hyperref}

\cvprfinalcopy 

\def\cvprPaperID{09479} 

\ifcvprfinal\fi
\begin{document}

\title{Learning to Observe: Approximating Human Perceptual Thresholds for Detection of Suprathreshold Image Transformations}

\author{
Alan Dolhasz, \hspace{0.6cm} Carlo Harvey, \hspace{0.6cm} Ian Williams
\and
Digital Media Technology Lab, Birmingham City University
\and
{\tt\small \{alan.dolhasz, carlo.harvey, ian.williams\}@bcu.ac.uk}
\and
{\tt\small \color{blue}{https://github.com/dmt-lab/learning-to-observe}}
}

\maketitle

\begin{abstract}
   Many tasks in computer vision are often calibrated and evaluated relative to human perception. In this paper, we propose to directly approximate the perceptual function performed by human observers completing a visual detection task. Specifically, we present a novel methodology for learning to detect image transformations visible to human observers through approximating perceptual thresholds. To do this, we carry out a subjective two-alternative forced-choice study to estimate perceptual thresholds of human observers detecting local exposure shifts in images. We then leverage transformation equivariant representation learning to overcome issues of limited perceptual data. This representation is then used to train a dense convolutional classifier capable of detecting local suprathreshold exposure shifts - a distortion common to image composites. In this context, our model can approximate perceptual thresholds with an average error of 0.1148 exposure stops between empirical and predicted thresholds. It can also be trained to detect a range of different local transformations. 
\end{abstract}

\section{Introduction}

Human observers are the target audience for image content and thus the ultimate judges of image quality, which is often measured with reference to opinions of humans and various \textit{local} and \textit{global} distortions and inconsistencies perceptible to them. These distortions can arise as side-effects of image acquisition, compression, transmission, compositing, and post-processing. Understanding and modeling how humans detect and process distortions to arrive at subjective quality scores underpin image quality assessment (IQA) research. Many attempts have been made at modeling the sensitivity of the human visual system (HVS) to certain types of distortions for applications primarily in IQA \cite{daly1992visible,  bradley1999wavelet, lai2000haar, wang2001designing, dusek2003testing, krupinski2004use} and saliency modeling \cite{van1996perceptual, wang2007video, lin2011perceptual, itti1998model}, where detection of relevant and perceptually suprathreshold features is key to the approximation of human performance. However, many of these approaches are limited in their generalizability, efficiency or transferability. Alternative approaches based on signal fidelity \cite{sheikh2006image}, statistical measures \cite{sheikh2005information} and deep learning models \cite{bosse2017deep, talebi2018nima} were also developed as a way to address such limitations.

\begin{figure}[t]
    \centering
    \includegraphics[width=1.0\linewidth, trim={13cm, 2cm, 5.5cm, 0cm},clip]{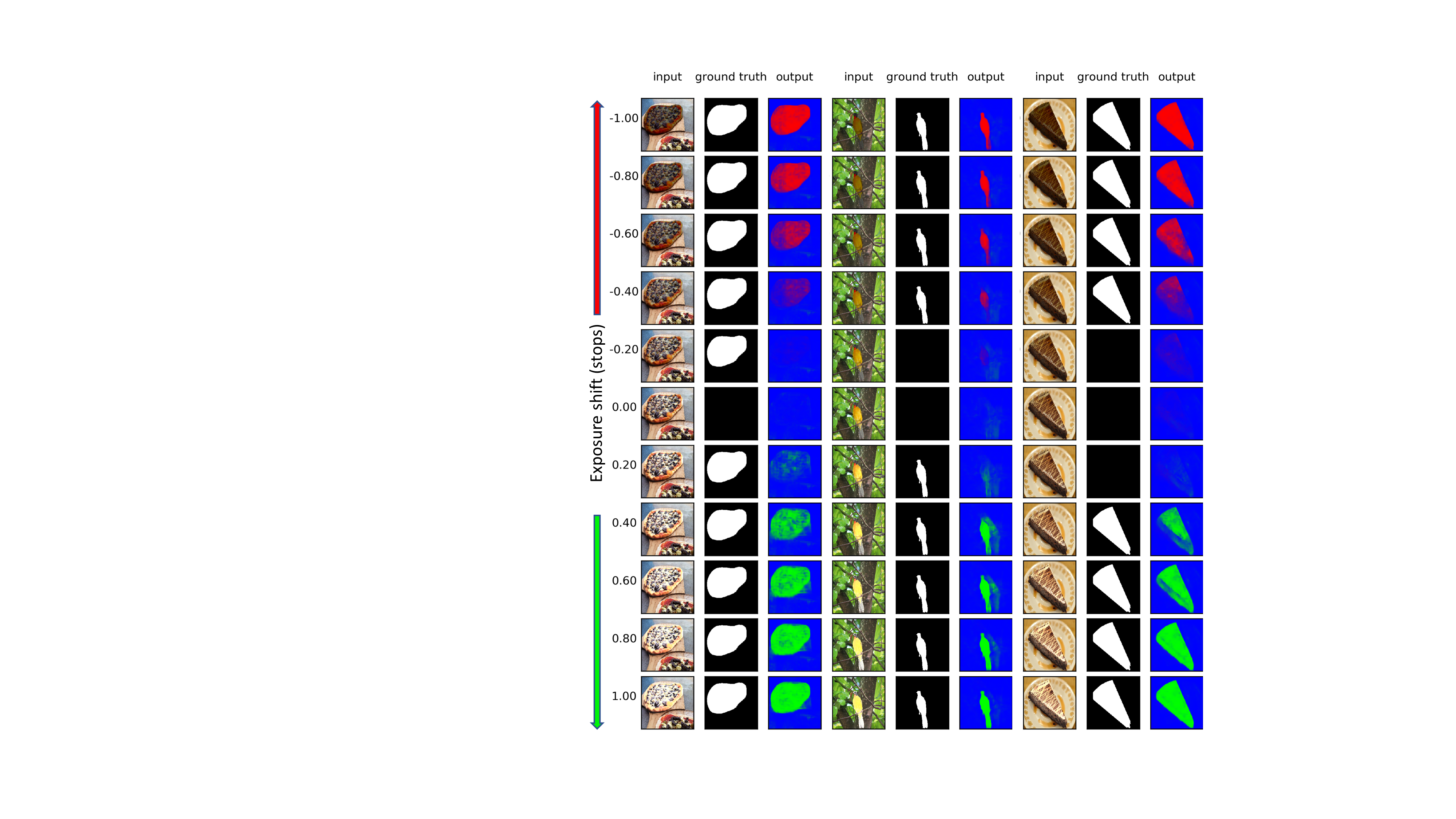}
    \caption{Performance of our model illustrated for three input images and 11 levels of exposure transformation. The left columns show input images with applied exposure transformations and the magnitude of this transformation expressed on a $log_2$ scale. Middle columns show ground truth from our subjective experiments and rightmost columns show output of our model, where {\color{red}red} and {\color{green}green} regions indicate detected negative and positive suprathreshold exposure transformations, while {\color{blue}blue} regions indicate no suprathreshold transformations.}
    \label{fig:example_system}
\end{figure}

Human sensitivity to physical stimuli is measured using psychophysics \cite{fechner1966elements} and often represented using psychometric functions, which describe observer performance as a function of stimulus intensity \cite{barten1999contrast}. This method is effective when stimuli are simple, but is difficult to generalize to more complex stimuli, such as natural images. This is largely due to the vast amount of variation in the set of natural images and the corresponding number of trials required to measure observer performance across sufficiently many images and stimulus intensities. 
In subjective image evaluation, the quality score can be seen as a result of applying an \textit{observer function} to an input image. This function can be summarized as detection of visible distortions, their implicit pooling, and mapping to a point on a given quality scale \cite{itu2002500}. This is further influenced by task, image content, and allocation of attention \cite{moorthy2009perceptually}. Recent work has made significant progress in approximating this entire process in the context of IQA using deep convolutional neural networks (DCNN) \cite{bosse2017deep, talebi2018nima}. However, these approaches are mostly limited to a fixed set of low level, globally-distributed artifacts available in public IQA datasets, such as LIVE \cite{sheikh2005live} containing 5 types of distortions, or TID2013 with 24 types and 5 magnitude levels each \cite{ponomarenko2013color}. This limits the generalizability, particularly for applications where the type and number of possible distortions vary significantly, or where the distortions are context-dependent and only present in a local region of the image, such as image compositing. The creation of such datasets is a costly and time-consuming process, due to the need for human observers. Approximation of this observer function - detecting visible inconsistencies of an arbitrary type - would allow for application in many areas related to IQA, including composite quality assessment, manipulation detection, and image restoration.

In this work, we propose a DCNN-based methodology to approximate this observer function and validate our method with respect to a specific local distortion common to image composites - \textit{local exposure inconsistencies} associated with an image region occupied by an object. We achieve this by learning a mapping between images affected by this distortion and corresponding points on an empirical psychometric function, estimated with respect to this distortion type. Viewing image distortions as transformations allows use of unsupervised methods for learning relevant features. Our approach can be applied to a range of problems where distortions visible to humans need to be localized in an image, such as IQA or composite quality assessment, even when little subjective data is available. Our contributions are:

\begin{itemize}[noitemsep,topsep=0pt,parsep=0pt,partopsep=0pt]
    \item A novel method for detecting effects of local image transformations based on perceptual data and unsupervised pre-training
    \item A model trained using this method to detect local exposure shifts
    \item A dataset of images with corresponding empirical subjective perceptual thresholds from our experiments
\end{itemize}

\section{Related Work}

\subsection{Human Perception}
The HVS displays different levels of sensitivity to various distortions and inconsistencies in images, detecting some readily \cite{biederman1982scene}, while disregarding others completely \cite{ostrovsky2005perceiving, cavanagh2005artist}. Detection of inconsistencies in lower-level properties of images depends largely on fundamental characteristics of the HVS, such as contrast sensitivity \cite{barten1999contrast}, luminance adaptation, and masking \cite{peli1990contrast}. These characteristics describe how immediate context, such as differences in background luminance, spatial frequency, and presence of texture, influence the visibility of different image artifacts. For example, distortions such as noise or quantization, are much easier to notice on a textureless background, compared to a textured one. The amount of change in stimulus required for an observer to reliably notice a difference is referred to as the a just-noticeable difference (JND) or difference limen. JNDs have been used extensively to model human perceptual sensitivity in tasks such as blur detection \cite{shi2015just}, visual attribute differences \cite{yu2015just}, perceptual metrics \cite{zhang2018unreasonable}, or 3D model attribute similarities \cite{cleju2006evaluation}. Observer sensitivity is further modulated by the allocation of visual attention \cite{ninassi2007does, liu2011visual}, particularly for localized distortions, such as those in image composites \cite{dolhasz2017poster}.

\subsection{Psychometric Functions}
Observers assessing image quality base their judgments on visual evidence, such as visible artifacts or distortions \cite{vu2008visual}. Human performance in detection and discrimination tasks is commonly modeled using psychometric functions  \cite{shi2015visual, scheirer2014perceptual, johnson2010using, wallis2012image, ninassi2006pseudo}. The psychometric function describes a relationship between observer performance and an independent variable, often describing a stimulus level or physical quantity \cite{wichmann2001psychometric}. It is defined as
\begin{equation}
    \Psi(x; \theta) = \gamma+(1-\gamma)f(x; \alpha, \beta)
\end{equation}

\noindent where $\theta$ refers to the set of parameters: $\gamma$ (guess rate) defines the lower bound of the function corresponding to chance performance, while $f(x; \alpha, \beta )$ defines a sigmoidal function parametrized by $\alpha$ - its location and $\beta$ - its slope. Observer performance for a given stimulus $x$ is represented by the output of $\Psi$ denoted as $y = \Psi(x; \theta)$. The threshold of a perceptual function can thus be defined as the stimulus level $x_t$ which yields a particular probability of stimulus detection $y_t$, such that $x_t = \Psi^{-1}(y_t)$. In practice, psychometric functions are commonly estimated using adaptive sampling procedures, such as QUEST \cite{watson1983quest}, which limit the number of required trials by sampling stimuli with the highest probability of lying at the threshold.

\subsection{Saliency \& Semantic Segmentation}
\label{sec:saliency}
Our work is related to both salient object detection (SOD) and semantic segmentation (SS), both of which seek to assign class membership of individual pixels based on local contextual information. SS assigns a single semantic object class to each pixel of an input image \cite{long2015fully}. SOD aims to segment the most salient object in an image, based on its low-level image-based features, often measured against human performance \cite{borji2015salient}. Image-to-Image neural networks have become popular tools in these domains, underpinning many state-of-the-art CNN architectures such as fully-convolutional networks (FCNs) \cite{chen2017deeplab}, U-nets \cite{ronneberger2015u}, adversarial approaches, such as Pix2Pix \cite{pix2pix2017} and many variations thereof. These approaches emphasise the importance of multi-scale features \cite{li2015visual}, as well as spatial resolution preservation through dilated convolution and skip connections \cite{yu2015multi, chen2014semantic}.

\subsection{Unsupervised \& Semi-Supervised Learning}
Supervised learning approaches, such as those in Section \ref{sec:saliency}, require large amounts of labeled data, which can necessitate a significant time effort. For perceptually-constrained tasks, this overhead is multiplied, due to the requirement for larger observer samples and more replications, compared to Likert-style subjective opinion studies. Conversely, unsupervised learning techniques do not require manually-labeled data to learn. Thus, this paradigm is attractive for our application, as we can exploit unlabelled data to learn the features describing a given transformation and then use a smaller, labeled perceptual dataset to fine-tune these features to the empirical perceptual data.

Some approaches, such as representation learning \cite{bengio2013representation}, relax the requirement for labeled data through the use of auto-encoders (AEs) and generative adversarial networks (GANs). AEs learn compressed representations of data by attempting to reconstruct it through a feature bottleneck. Representations learned by AEs tend to encode salient features of the data they are conditioned on, which in turn can act as a task-specific feature extractor for supervised tasks \cite{baldi2012autoencoders}. On the other hand, GANs adopt an adversarial training regime, where a generator and discriminator are jointly trained. E.g. the generator can be tasked with generating a sufficiently realistic image, such that the discriminator classifies it as real. In turn, the discriminator is tasked with separating generated images from real ones \cite{radford2015unsupervised}. Zhang et al. (2019) showed that the performance of supervised classifiers can be improved by using an Auto-Encoding Transformations paradigm. They propose to learn transformation equivariant representations (TERs), which encode transformations applied to the input \cite{zhang2019aet}. This reduces the need for data augmentation and forces the encoder to learn a better representation of the input data, which encodes visual structures well, invariant of the transformation of the input. We adapt this approach to detecting local transformations within an image, which forms the foundation of our proposed methodology.

\section{Method}

In this section, we elaborate on our proposed approach and detail our model design and rationale. We summarize our methodology, including the formulation of distortions as transformations, use of empirical perceptual thresholds as decision boundaries, collection of empirical psychometric data, training dataset preparation, and both stages of our training procedure.

\subsection{Distortions as Transformations}


Many distortions affecting image quality can be seen as transformations applied to the original, uncorrupted image as a side-effect of some processes such as transmission, compositing, or compression. This is conceptually similar to the intuition behind denoising autoencoders \cite{bengio2017deep}. Denoising autoencoders learn a low-dimensional manifold near which training data concentrate. They also implicitly learn a function projecting corrupted images $\Tilde{I}$, affected by a corruption process and lying near the manifold of uncorrupted images, back onto this manifold. This conceptualization allows for the generation of large amounts of training data from a small set of undistorted images, by applying various transformations. We focus on a single transformation: local exposure shifts. This corresponds to the scaling of luminance by a constant, applied to a region within image $I$ corresponding to an object and defined by a binary mask $M$. This is performed on the luminance channel of the perceptually-uniform $Lab$ colorspace \cite{robertson1977cie}. We motivate this choice as follows: observers are reliable at detecting such low-level image distortions \cite{dolhasz2016measuring}; exposure distortions represent common mismatches present in image composites, which are a motivating application of our research \cite{xue2012understanding}; this type of transformation is computationally inexpensive to apply, allowing for gains in training efficiency.

\subsection{Perceptual Thresholds as Decision Boundaries}

In the context of image distortions and assuming controlled viewing conditions, a psychometric function can be seen as the result of an observer process operating on a range of input data. Given an unprocessed image $I$, object mask $M$, observer function $O$ and $\Tilde{I}_x$ a corrupted version of $I$ resulting from a local transformation $T(I,M,x)$, the empirical psychometric function can be interpreted as a result of applying the observer function to $\Tilde{I}$ for all values of $x$. The observer function $O$ thus represents the perceptual process performed by an observer, which maps an input stimulus $\Tilde{I}_x$ to a point on the psychometric function. Accordingly, detecting suprathreshold transformations in an image can be defined as applying the observer model to classify each pixel based on the existence of the effects of a suprathreshold transformation. This requires \textbf{a)} a psychometric function describing observer performance with respect to the magnitude of the transformation and specific image stimulus, \textbf{b)} contextual information about the scene and appearance of objects within it, from which information about the existence of local distortions can be derived and \textbf{c)} an appropriate feature representation, equivariant to the transformation in the training data. Consequently, our problem can be defined as a \textit{pixel-wise classification} of an image, where each pixel is assigned one of three classes $c$, whose decision boundaries are defined by the thresholds $x_{t-}$ and $x_{t+}$ of the two psychometric functions estimated for a given image, with respect to the parameter $x$ of the transformation generating the stimuli $\Tilde{I}$:

\begin{equation}
c=
    \begin{cases}
    0,              & \text{if } x < x_{t-}\\
    1,              & \text{if } x > x_{t+}\\
    2,              & \text{otherwise}
\end{cases}
\label{eq:classes}
\end{equation}

\begin{figure}
    \centering
    \includegraphics[width=1.0\linewidth, trim={0cm, 4cm, 14cm, 0cm}, clip]{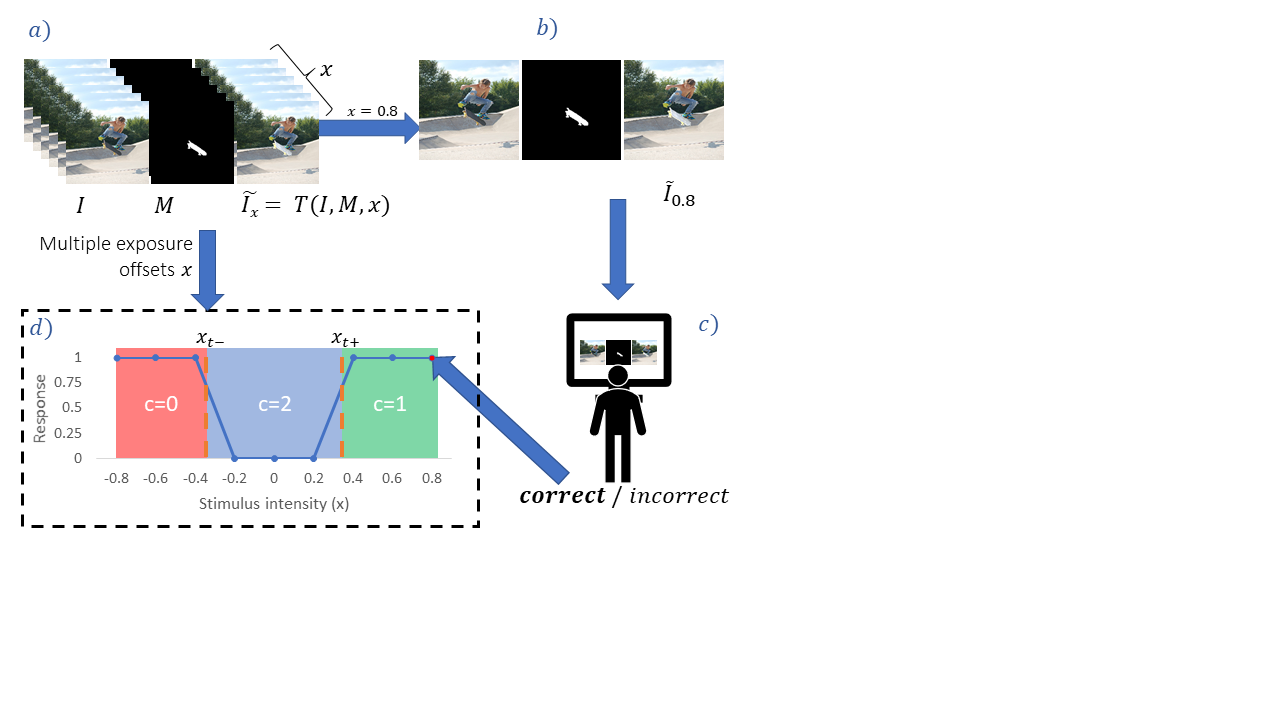}
    \caption{Illustration of the 2AFC procedure used in our experiments. a) For a given image $I$ and object mask $M$ we generate images $\Tilde{I}$ with different exposure offsets based on the sampled value of $x$. b) Example stimulus displayed to an observer. c) Observer correctly identifies $I$ and $\Tilde{I}$ for $x=0.8$. d) Observer response added to their previous responses for different sampled values of $x$. Symbols $x_{t-}$ and $x_{t-}$, illustrated with orange dashed lines, indicate the location of the threshold after performing psychometric function fitting.}
    \label{fig:overview_slide}
\end{figure}

\noindent Here, $x_{t}$ is the value of the transformation parameter for which the probability of detection exceeds threshold $t$, set to $0.75$, corresponding to the JND in 2AFC tasks. This is the midpoint between perfect (100\%) and chance (50\% for 2AFC task) performance \cite{wichmann2001psychometric}. As we capture two psychometric functions per image, one corresponding to decreasing the pixel intensity ($x_{t-}$) and one for increasing it ($x_{t+}$), their two thresholds separate the parameter space $x$ into three regions (Fig. \ref{fig:overview_slide}d).

\subsection{Psychometric Function Estimation}

To estimate image-wise empirical psychometric functions with respect to our transformation, we design a 2AFC study using a dataset of natural images with segmented objects, where the segmentation is defined by a binary mask. Following the approach of \cite{dolhasz2016measuring}, we systematically apply transformations with different values of $x$ to the segmented object. We display the \textit{original} ($I$) and \textit{transformed} ($\Tilde{I}$) images side by side in random order and ask observers to identify $I$ correctly. We repeat this for multiple values of $x$ and fit Weibull psychometric functions to each observer's responses for each image. To extract the thresholds, we estimate the parameter values $x_{t-}$ and $x_{t+}$ corresponding to a performance level of $y_{t}$ for negative and positive exposure shifts, respectively. We then bootstrap mean thresholds across all observers who viewed the same image. We detail the stages of this process in the remainder of this section.

\subsubsection{Experiment Design}
All experiments are performed under controlled laboratory conditions, following the ITU BT-500 recommendation \cite{itu2002500}. We use an Apple Cinema HD 23" monitor, calibrated to sRGB colorspace using an X-Rite i1Display Pro display calibration device. Observers are positioned 65cm away from the display. To mitigate the confounding impact of visual search on the task, particularly when differences between the images are minimal, we explicitly indicate the transformed region in the image by displaying the binary mask corresponding to the object, following \cite{dolhasz2017poster}. To minimize the number of experimental trials we leverage the QUEST adaptive sampling procedure \cite{watson1983quest}, using the implementation from the PsychoPy 2 library \cite{peirce2019psychopy2}. 

\subsubsection{Observers \& Stimuli}
We recruit $N\!\!=\!\!120$ naive observers, with a mean age of $31$ $(SD=11.85)$, $44$ of whom are female and randomly assign them to 20 groups. Observers are screened for normal vision before participating in the experiment. Our stimuli dataset consists of 300 8-bit images with corresponding object masks, randomly sampled from the LabelMe \cite{russell2008labelme} and SUN \cite{xiao2010sun} datasets. These images are then evenly distributed across the observer groups. Each group views 15 unique images from the dataset.

\subsubsection{Task \& Experimental Procedure}
In the experimental session, each observer performs repeated 2AFC trials for each of the 15 base images in their allocated image sample, viewing at least 20 different variations of each base image. Observers first complete 20 trials using a calibrating image, results for which are discarded. In each trial observers are shown 2 images: the original image $I$ and a transformed version of the original image $\Tilde{I}_x$, the result of exposure transformation $T(I,M,x)$ of magnitude $x$. A segmentation mask $M$ is also displayed indicating the target object. These images are displayed at the same time and remain on-screen for 5 seconds. The order of $I$ and $\Tilde{I}$ is randomized every trial. Observers are instructed to correctly indicate $I$ by clicking a corresponding button. After each response, a new value of $x$ is sampled by the QUEST procedure \cite{watson1983quest}, and the process is repeated 20 times. 

\subsubsection{Perceptual Threshold Estimation}
For each observer-image combination, we collect binary responses $y$ with corresponding stimulus intensities $x$. We use the PsychoPy library \cite{peirce2019psychopy2} to fit a Weibull cumulative distribution function to this data, given by
\begin{equation}
    y = 1 - (1-\gamma)\scalebox{1.2}{$e^{-(\frac{k x}{t})^\beta}$}
\end{equation}
\noindent and 
\begin{equation}
    k = -log\bigg(\frac{1-\alpha}{1-\gamma}\bigg)\scalebox{1.6}{$^\frac{1}{\beta}$}
\end{equation}

\noindent where $x$ is the stimulus intensity, $y$ is the proportion of correct responses, $\gamma$ is the performance level expected at chance, equal to 0.5 for 2AFC tasks, $\alpha$ is the performance level defining the threshold (set to 0.75, corresponding to the JND for 2AFC), $\beta$ is the slope of the function and $t$ is the threshold. Once we extract the threshold of this function, we pool the threshold values across observers for that image and bootstrap the mean of these thresholds, using $1000$ bootstrap samples. We obtain two generalized perceptual thresholds: $x_{t-}$ and $x_{t+}$ for each image in our dataset.

\begin{figure*}[t]
\begin{center}
\includegraphics[width=0.90\textwidth,trim={1cm 3.0cm 2cm 2.0cm},clip]{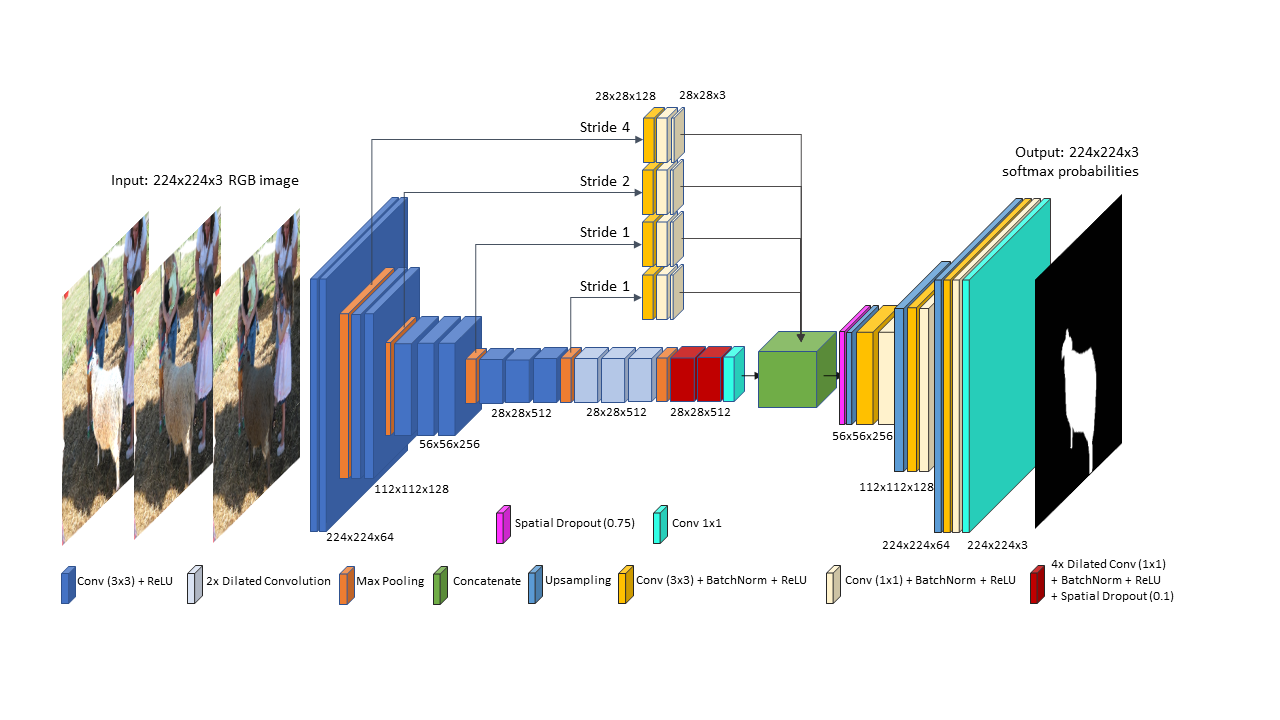}
\end{center}
\caption{Architecture of our VGG16-based convolutional autoencoder used in the perceptual threshold learning task. The network is based on a FCN adaptation of the VGG16. See Section \ref{sec:TER} for a detailed description of the architecture.}
\label{fig:short}
\end{figure*}

\subsection{Transformation Equivariant Representation Learning (AET)}
\label{sec:TER}

While object classifiers, such as models trained on ImageNet, aim to achieve invariance to changes in object brightness, our task explicitly uses these features to assign classes to output pixels. Thus, transfer learning with an object classifier/detector is unsuitable for addressing overfitting with our small dataset. Instead, we propose to first learn a task-specific TER in an unsupervised manner, adopting the AET approach of Zhang et al. \cite{zhang2019aet}, who encode a TER by training to predict transformation parameters that describe a transformation between two inputs. Analogously, we wish to encode a representation that is invariant to a particular transformation type: local exposure shifts.

\subsubsection{AET: Network Architecture}
We can train a convolutional autoencoder to predict the parameter of a local exposure shift applied to the input, by mapping images containing local exposure shifts to masks indicating their pixel-wise magnitude. To achieve this, we develop an AET model based on the VGG16. We first convert the VGG16 to a fully convolutional network \cite{long2015fully}. Due to the importance of contextual and multiscale information to our task, we attach a multiscale extension, as proposed in \cite{li2018contrast}. This introduces skip connections to the model, taking outputs after each max pooling layer in the VGG16 and passing each through an additional convolutional branch before concatenating the output of all branches. Each branch consists of 3 convolutional blocks. The first block contains a $3\times3$, 128-channel convolutional layer with a stride setting dependant on the scale of the input. This is $4$, $2$, $1$, $1$ respectively for inputs from the first 4 max pooling layers, causing all multiscale branches outputting feature maps of equal resolution. This layer is followed by a batch normalization layer and a ReLU activation. The following two blocks contain $1\times1$ convolutional layers with a stride of $1$, with $128$ and $3$ channels respectively. They are each followed by batch normalization and a ReLU activation. To output masks of equal resolution to the input images, we add a convolutional decoder to the output of the multiscale concatenation layer in our model. It consists of 3 blocks, each block containing a $2\times$ upsampling layer, followed by two sets of convolution, batch normalization, and ReLU layers. The first convolution in the block uses $3\times3$ kernels, while the second uses $1\times1$ kernels. See Figure \ref{fig:short} for a detailed overview.

Using this architecture, we design an AET model which shares the weights of the network between two image inputs, $I$ and $\Tilde{I}_{x}$ (Fig. \ref{fig:aet}). Activations for both inputs are concatenated and fed to a final convolutional layer. As our transformation can be expressed by a single scalar the final layer of our AET is a $3\times3$ convolutional layer with a linear activation, which outputs masks with resolution equal to the input image, with a single value expressing the predicted exposure shift for each pixel. This way we can train our model to approximate pixel-wise transformations applied to an input image.

\subsubsection{AET: Training Data Generation}
To train the AET in an unsupervised manner, we learn a mapping between input images $\Tilde{I}$ and output masks $Y = xM$, which encode the parameter of the transformation applied to the input. $\Tilde{I}$ contains an exposure shift applied within the region defined by $M$. Each pixel in $Y$ contains the value of the exposure shift $x$ applied to the corresponding pixel in $\Tilde{I}$. This is $x$ wherever $M=1$ and $0$ elsewhere (Fig. \ref{fig:aet}). During training, we dynamically sample images $I$ and corresponding masks $M$ from the MSCOCO dataset \cite{lin2014microsoft}. As some images in MSCOCO contain multiple masks, we randomly select one of them, provided its area is larger than 1\%. We then apply exposure shifts by sampling the transformation parameter $x$ and scaling the luminance channel of $I$ after conversion to $Lab$ colorspace:

\begin{figure}
    \centering
    \includegraphics[width=\linewidth, trim={0.5cm, 7cm, 9cm, 0cm},clip]{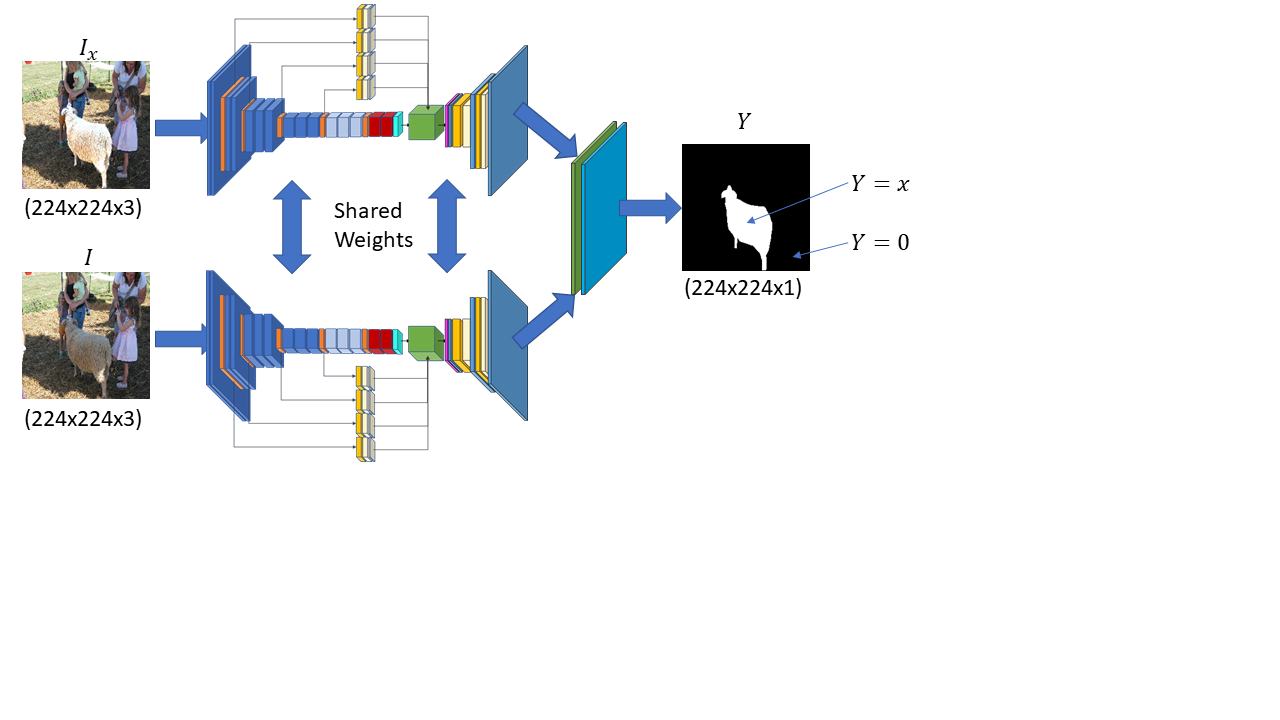}
    \caption{Unsupervised AET architecture consisting of a VGG16-based convolutional autoencoder with weights shared across two inputs. Activations for both inputs are then concatenated and fed to a final convolutional layer with a single channel output. The output masks encode the parameter of the transformation for each pixel.}
    \label{fig:aet}
    \vspace{-0.5cm}
\end{figure}

\begin{equation}
    \Tilde{I_L} = 2^{x} I_L \odot M + I_L \odot (1-M)
    \label{eq:hadamard}
\end{equation}

\noindent where $x$ is a scalar sampled from a base-2 log-uniform distribution spanning $(log_2(0.1),$ $log_2(10))$, $I_L$ is the luminance channel of the original image $I$ after conversion from $RGB$ to $Lab$ colorspace, $M$ is the alpha mask and $\odot$ is the Hadamard product. We clip the pixel values of processed image to the range $(0.0, 1.0)$, convert back to $RGB$, rescale to $0.0$ mean and unit variance, reshape images to $(224,224,3)$ and feed both $I$ and $\Tilde{I}$ to the two inputs of the AET (as in Fig. \ref{fig:aet}). The output of the network is a mask $\hat{Y}$ approximating the parameter of the transformation at each pixel of the input image. 

\subsubsection{AET: Objective \& Optimizer Details}
\label{sec:optimization}
We train our model using the Adam optimizer \cite{kingma2014adam}. We use default values for all parameters, aside from the learning rate, which is controlled using a cosine annealing schedule \cite{loshchilov2016sgdr}. The minimum and maximum learning rate in the annealing schedule are set to $1\mathrm{e}{-6}$ and $1\mathrm{e}{-4}$, respectively. The learning rate cycles between these values over 5 epochs, after which the maximum learning rate is reduced to 90\% of its value, and the cycle is repeated for $1.5\times$ as many epochs. We train the AET for 90 epochs, minimizing the mean squared error (MSE) loss between $\hat{Y}$ and $Y$. We use the model with the lowest validation error as the backbone for the Perceptual Threshold Classifier.

\subsection{Perceptual Threshold Classifier (PTC)}
\label{sec:threshold_classifier}
\subsubsection{PTC: Network Architecture}
To detect perceptually suprathreshold transformations in images, we utilize the pre-trained AET architecture described in Section \ref{sec:TER}, extract the encoder and decoder shown in Figure \ref{fig:short} and replace the final single-channel convolutional layer of the decoder with a spatial dropout layer with a dropout probability of 75\%, followed by a 3-channel convolutional layer with a softmax activation.

\subsubsection{PTC: Training Data Generation}
Using thresholds obtained in our experiments, we devise a data generation method which dynamically applies random exposure transformations to the images used in our 2AFC experiment and generates corresponding categorical masks, based on whether the parameter of the transformation $x$ exceeds one of the two empirical thresholds estimated for a given image. When $x$ exceeds a threshold, any pixels affected by this suprathreshold transformation are assigned $c=0$ (negative suprathreshold exposure shift) or $c=1$ (positive suprathreshold exposure shift), following Equation \ref{eq:classes}. The last channel of the target image corresponding to $c=2$ is conceptually similar to the \textit{background} class in semantic segmentation models. It indicates pixels that do not belong to any of the foreground classes. In our case, these are pixels unaffected by a suprathreshold transformation. We use a 90\%-10\% training/validation split. The shape of the target mask is $(224,224,3)$, containing one channel per class. During training, we use a data generator constrained to ensure a balanced class distribution in each minibatch. Specifically, for each batch we sample $x$ from three random distributions whose ranges are defined by the perceptual thresholds for a given image:

\begin{equation}
x \in \mathbb{R}:
    \begin{cases}
    (log_2(0.1), x_{t-}),  & \text{if } x < x_{t-}\\
    (x_{t+}, log_2(10)),   & \text{if } x > x_{t+}\\
    [x_{t-}, x_{t+}],     & \text{otherwise}
\end{cases}
\label{eq:classes_sampling}  
\end{equation}

The distribution for $c=2$ is log-uniform, whereas the distributions for classes 0 and 1 are exponential distributions biased towards values of $x$ lying close to the thresholds $x_{t-}$ and $x_{t+}$ respectively. These three values of $x$ are then used to create three processed images and corresponding target masks $Y$, one for each class. For larger batch sizes we simply sample multiple images for each class. To improve generalization, we apply image augmentation, limiting to zooming, rotation, and cropping in order not to affect relative pixel intensities. We perform horizontal and vertical flipping with 50\% probability, as well as random scaling and cropping in the range 110-150\% and with 50\% probability.

\subsubsection{PTC: Objective \& Optimizer Details}
We follow the optimization approach from Section \ref{sec:optimization} with minor changes. Firstly, we select a loss function appropriate for pixel-wise classification with an imbalanced dataset. In most images in our dataset the background class occupies more pixels than either of the suprathreshold classes, we handle this imbalance by reducing the contribution of easy classification examples to the loss using focal loss \cite{lin2017focal}. We also experiment with freezing different sections of our backbone network in order to maximize generalizability. We train our models with a batch size of 12 until convergence using early stopping to cease training when no improvement in validation loss is seen for 400 epochs. For evaluation, we select the model which maximizes the validation mean intersection-over-union measure.

\section{Results \& Discussion}
\begin{figure}[b]
    \centering
    \includegraphics[width=1.0\linewidth, trim={1cm 0cm 1cm 1cm}, clip]{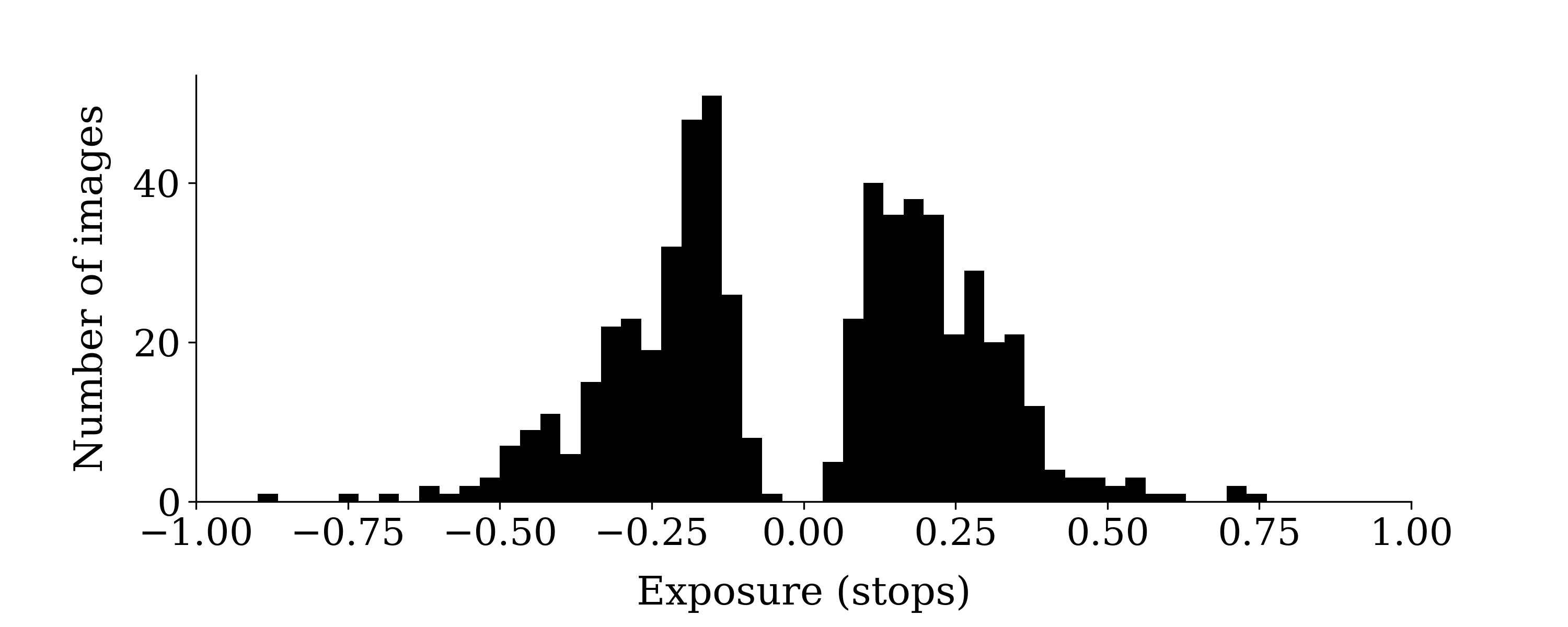}
    \caption{Empirical thresholds collected in our experiment}
    \label{fig:thresholds}
\end{figure}

\begin{figure*}[ht]
    \centering
    \includegraphics[width=1.0\textwidth, trim={4cm 1.8cm 5cm 0cm}, clip]{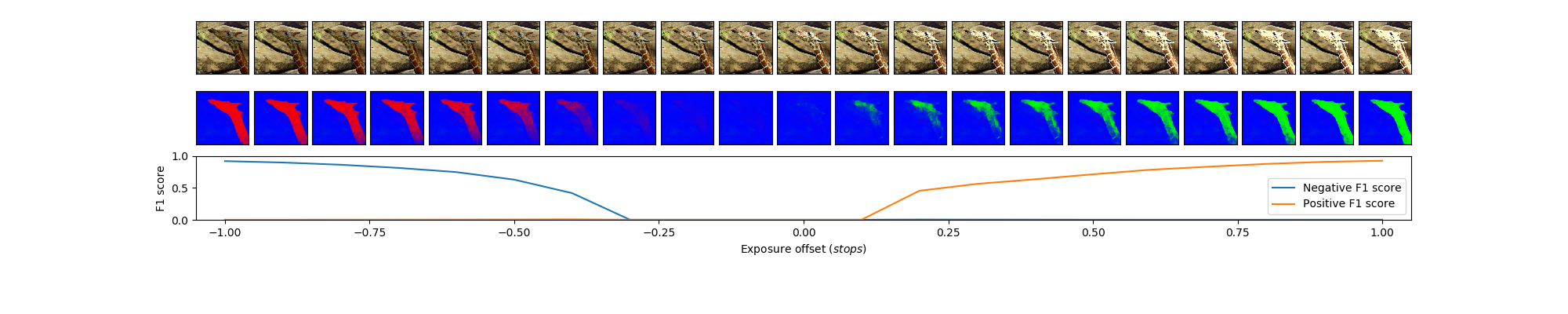}
    \caption{Illustration of how change in $F1$ score between predicted and ground truth (not shown here) masks is used to estimate our model's decision boundary. The top row shows input images, the middle row shows model prediction softmax probabilities with {\color{red} red} for detected negative offsets (class 0), {\color{green} green} for positive offsets (class 1) and {\color{blue} blue} for no offset. The bottom row shows class-wise $F1$ scores for classes 0 and 1. More examples can be found in supplementary materials.}
    \label{fig:f1_vis}
\end{figure*}

\subsection{Perceptual Threshold Estimation}
In our 2AFC study, we obtained a total of $41725$ unique responses, with an average of $23.14$ responses per observer per image. Observers took on average $2.65s$ per response. A total of $590$ mean thresholds for $295$ images were calculated after fitting psychometric functions, bootstrapping and removing outlier thresholds beyond 3 standard deviations (Fig. \ref{fig:thresholds}). The means of the resulting threshold distributions were $x_{t-}=-0.2478$ and $x_{t+}=0.2280$ for negative and positive thresholds respectively. On average, perceptual thresholds were lower for highly-textured and bright objects. We found significant correlations between the mean luminance of target objects and corresponding mean thresholds. For negative offsets the Pearson product-moment correlation coefficient was $r=.25$ $p\leq.001$ and $r=-.39$ $p\leq.001$ for positive offsets. We found a similar correlation between the standard deviation of object luminance values: $r=.30$ $p\leq.001$ for negative and $r=-.45$ $p\leq.001$ for positive offsets. No significant correlations between perceptual thresholds and target object areas were observed. However, we note that the highest perceptual thresholds in our results were observed in images with very small objects. In post-test discussions, observers reported selecting specific parts of objects to inform their decisions, these were commonly high-contrast regions near target object boundaries.

\subsection{Perceptual Threshold Learning}

Since no previous work has addressed the problem of perceptual threshold approximation, we cannot compare our model's performance to existing solutions. Instead to evaluate the validity of our approach we perform 5-fold cross-validation, reporting average MSE between the predicted and ground truth thresholds for our validation set. We first develop a psychometrics-inspired method for finding our model's decision boundary, which will serve as a threshold to be compared against empirical thresholds from our experiments. This is done by calculating the soft $F1$ score for each of the two suprathreshold classes between the ground truth mask and model prediction for a range of values of $x$ and placing a threshold at the point when $F1$ score becomes nonzero. In our experiments we use $F1=0.1$, see Figure \ref{fig:f1_vis} for an illustration of the soft F1 score as a function of exposure shift. More visual examples can be found in the supplementary materials.

To evaluate the relevance of features learned by the AET, we perform this analysis for a range of fine-tuning regimes, where different parts of the model are frozen before training. The results of this experiment can be seen in Table \ref{tab:freeze}. Overall, our results indicate the benefits of adopting both the AET and multiscale extension, particularly considering the performance increase afforded by freezing the entire encoder and only fine-tuning the decoder. The model's performance drops significantly when the pre-training stage is omitted or when all layers of the pre-trained model are allowed to be fine-tuned.  

\begin{figure}
    \centering
    \includegraphics[width=0.85\linewidth, trim={0cm 12cm 15.5cm 0cm}, clip]{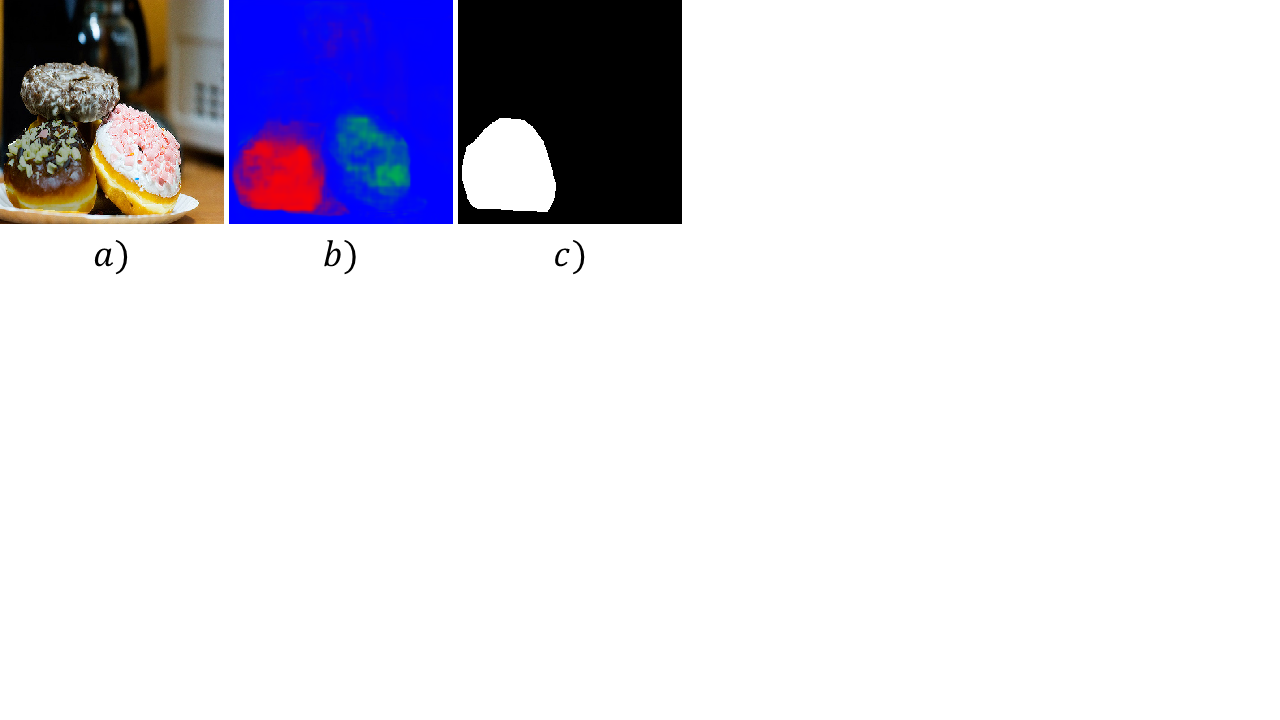}
    \caption{Example of \textit{a)} Over-exposure resulting from flash or spot lighting in the original image \textit{b)} both the original over-exposure {\color{green}green}) and manually applied underexposure ({\color{red}red}) are detected by our model \textit{c)} mask showing area where negative exposure shift is manually applied}
    \label{fig:donut}
\end{figure}
\begin{table}[]
    \centering
    \resizebox{1.0\linewidth}{!}{
    \begin{tabular}{lrrr}
        \toprule
        Freeze Up To Layer & MSE both & MSE $x_{t-}$ & MSE $x_{t+}$ \\ \midrule
        no freeze & 3.9690 & 3.5716 & 4.3664 \\ 
        block1 pool & 0.3028 & 0.2618 & 0.3442 \\
        block2 pool & 0.2098 & 0.2188 & 0.2000 \\
        block3 pool & 0.1895 & 0.1633 & 0.2161 \\
        block4 pool & 0.2350 & 0.2025 & 0.2681 \\
        block5 pool & 0.1335 & 0.1624 & 0.1046 \\
        concatenate & \textbf{0.1148} & \textbf{0.1307} & \textbf{0.0978} \\ \bottomrule \\
    \end{tabular}
    }
    \caption{Cross-validation results: Average mean squared validation errors between ground truth thresholds and model predictions are given in exposure stops. Individual errors for positive and negative exposure offsets are shown in the rightmost two columns. Errors in each row are a result of freezing progressive parts of the pre-trained AET backbone.}
    \label{tab:freeze}
\end{table}
\section{Conclusions, Limitations and Future Work}
We have presented a novel methodology for the detection of local suprathreshold image transformations based on approximating the function performed by an observer. This is achieved by training a fully convolutional image classifier and conditioning its class decision boundaries using a data generation scheme based on empirical perceptual thresholds corresponding to JNDs. We find that the threshold distributions generated by our model approximate the empirical threshold distributions from our experiments. We also confirm that adopting the unsupervised AET approach achieves consistently lower errors than training directly on the empirical data without pre-training. Our method can be applied to a range of local distortions or transformations, such as color shifts, blur, aliasing or subsampling, as long as they can be represented by a transformation and mask. Aside from transformations applied manually, our model detects pre-existing over-exposure in our validation set (see Fig. \ref{fig:donut}). Our results are constrained by the 8-bit dynamic range of images used in our study and the inherent biases associated with individual observers. However, they show that using CNN architectures and an AET unsupervised pre-training strategy if an efficient method of detecting local transformations in images. While a further detailed study and fine-grained optimization are required to maximize performance, our methodology is effective at approximating perceptual thresholds with respect to a local image transformation.
We are currently performing an extended study of our approach against different backbone architectures, training regimes, and optimization strategies. We also intend to apply our methodology as the first stage in automatic composite quality improvement. 

{\small
\bibliographystyle{ieee_fullname}
\bibliography{egbib}

\begin{thebibliography}{10}\itemsep=-1pt

\bibitem{baldi2012autoencoders}
Pierre Baldi.
\newblock Autoencoders, unsupervised learning, and deep architectures.
\newblock In {\em Proceedings of ICML workshop on unsupervised and transfer
  learning}, pages 37--49, 2012.

\bibitem{barten1999contrast}
Peter~GJ Barten.
\newblock {\em Contrast sensitivity of the human eye and its effects on image
  quality}, volume~21.
\newblock Spie optical engineering press Bellingham, WA, 1999.

\bibitem{bengio2013representation}
Yoshua Bengio, Aaron Courville, and Pascal Vincent.
\newblock Representation learning: A review and new perspectives.
\newblock {\em IEEE transactions on pattern analysis and machine intelligence},
  35(8):1798--1828, 2013.

\bibitem{bengio2017deep}
Yoshua Bengio, Ian Goodfellow, and Aaron Courville.
\newblock {\em Deep learning}, volume~1.
\newblock Citeseer, 2017.

\bibitem{biederman1982scene}
Irving Biederman, Robert~J Mezzanotte, and Jan~C Rabinowitz.
\newblock Scene perception: Detecting and judging objects undergoing relational
  violations.
\newblock {\em Cognitive psychology}, 14(2):143--177, 1982.

\bibitem{borji2015salient}
Ali Borji, Ming-Ming Cheng, Huaizu Jiang, and Jia Li.
\newblock Salient object detection: A benchmark.
\newblock {\em IEEE transactions on image processing}, 24(12):5706--5722, 2015.

\bibitem{bosse2017deep}
Sebastian Bosse, Dominique Maniry, Klaus-Robert M{\"u}ller, Thomas Wiegand, and
  Wojciech Samek.
\newblock Deep neural networks for no-reference and full-reference image
  quality assessment.
\newblock {\em IEEE Transactions on Image Processing}, 27(1):206--219, 2017.

\bibitem{bradley1999wavelet}
Andrew~P Bradley.
\newblock A wavelet visible difference predictor.
\newblock {\em IEEE Transactions on Image Processing}, 8(5):717--730, 1999.

\bibitem{cavanagh2005artist}
Patrick Cavanagh.
\newblock The artist as neuroscientist.
\newblock {\em Nature}, 434(7031):301, 2005.

\bibitem{chen2014semantic}
Liang-Chieh Chen, George Papandreou, Iasonas Kokkinos, Kevin Murphy, and Alan~L
  Yuille.
\newblock Semantic image segmentation with deep convolutional nets and fully
  connected crfs.
\newblock {\em arXiv preprint arXiv:1412.7062}, 2014.

\bibitem{chen2017deeplab}
Liang-Chieh Chen, George Papandreou, Iasonas Kokkinos, Kevin Murphy, and Alan~L
  Yuille.
\newblock Deeplab: Semantic image segmentation with deep convolutional nets,
  atrous convolution, and fully connected crfs.
\newblock {\em IEEE transactions on pattern analysis and machine intelligence},
  40(4):834--848, 2017.

\bibitem{cleju2006evaluation}
Ioan Cleju and Dietmar Saupe.
\newblock Evaluation of supra-threshold perceptual metrics for 3d models.
\newblock In {\em Proceedings of the 3rd symposium on Applied perception in
  graphics and visualization}, pages 41--44, 2006.

\bibitem{daly1992visible}
Scott~J Daly.
\newblock Visible differences predictor: an algorithm for the assessment of
  image fidelity.
\newblock In {\em Human Vision, Visual Processing, and Digital Display III},
  volume 1666, pages 2--15. International Society for Optics and Photonics,
  1992.

\bibitem{dolhasz2017poster}
Alan Dolhasz, Maite Frutos-Pascual, and Ian Williams.
\newblock [poster] composite realism: Effects of object knowledge and
  mismatched feature type on observer gaze and subjective quality.
\newblock In {\em 2017 IEEE International Symposium on Mixed and Augmented
  Reality (ISMAR-Adjunct)}, pages 9--14. IEEE, 2017.

\bibitem{dolhasz2016measuring}
Alan Dolhasz, Ian Williams, and Maite Frutos-Pascual.
\newblock Measuring observer response to object-scene disparity in composites.
\newblock In {\em 2016 IEEE International Symposium on Mixed and Augmented
  Reality (ISMAR-Adjunct)}, pages 13--18. IEEE, 2016.

\bibitem{dusek2003testing}
Jaroslav Dusek and Karel Roub{\'\i}k.
\newblock Testing of new models of the human visual system for image quality
  evaluation.
\newblock In {\em Seventh International Symposium on Signal Processing and Its
  Applications, 2003. Proceedings.}, volume~2, pages 621--622. IEEE, 2003.

\bibitem{fechner1966elements}
Gustav~Theodor Fechner, Davis~H Howes, and Edwin~Garrigues Boring.
\newblock {\em Elements of psychophysics}, volume~1.
\newblock Holt, Rinehart and Winston New York, 1966.

\bibitem{pix2pix2017}
Phillip Isola, Jun-Yan Zhu, Tinghui Zhou, and Alexei~A Efros.
\newblock Image-to-image translation with conditional adversarial networks.
\newblock {\em CVPR}, 2017.

\bibitem{itti1998model}
Laurent Itti, Christof Koch, and Ernst Niebur.
\newblock A model of saliency-based visual attention for rapid scene analysis.
\newblock {\em IEEE Transactions on Pattern Analysis \& Machine Intelligence},
  (11):1254--1259, 1998.

\bibitem{itu2002500}
Recommendation ITU-R~BT.
\newblock 500-11, methodology for the subjective assessment of the quality of
  television pictures”.
\newblock {\em International Telecommunication Union, Tech. Rep}, 2002.

\bibitem{johnson2010using}
Jeffrey~P Johnson, Elizabeth~A Krupinski, Michelle Yan, Hans Roehrig, Anna~R
  Graham, and Ronald~S Weinstein.
\newblock Using a visual discrimination model for the detection of compression
  artifacts in virtual pathology images.
\newblock {\em IEEE transactions on medical imaging}, 30(2):306--314, 2010.

\bibitem{kingma2014adam}
Diederik~P Kingma and Jimmy Ba.
\newblock Adam: A method for stochastic optimization.
\newblock {\em arXiv preprint arXiv:1412.6980}, 2014.

\bibitem{krupinski2004use}
Elizabeth~A Krupinski, Jeffrey Johnson, Hans Roehrig, John Nafziger, Jiahua
  Fan, and Jeffery Lubin.
\newblock Use of a human visual system model to predict observer performance
  with crt vs lcd display of images.
\newblock {\em Journal of Digital Imaging}, 17(4):258--263, 2004.

\bibitem{lai2000haar}
Yung-Kai Lai and C-C~Jay Kuo.
\newblock A haar wavelet approach to compressed image quality measurement.
\newblock {\em Journal of Visual Communication and Image Representation},
  11(1):17--40, 2000.

\bibitem{li2015visual}
Guanbin Li and Yizhou Yu.
\newblock Visual saliency based on multiscale deep features.
\newblock In {\em Proceedings of the IEEE conference on computer vision and
  pattern recognition}, pages 5455--5463, 2015.

\bibitem{li2018contrast}
Guanbin Li and Yizhou Yu.
\newblock Contrast-oriented deep neural networks for salient object detection.
\newblock {\em IEEE transactions on neural networks and learning systems},
  29(12):6038--6051, 2018.

\bibitem{lin2017focal}
Tsung-Yi Lin, Priya Goyal, Ross Girshick, Kaiming He, and Piotr Doll{\'a}r.
\newblock Focal loss for dense object detection.
\newblock In {\em Proceedings of the IEEE international conference on computer
  vision}, pages 2980--2988, 2017.

\bibitem{lin2014microsoft}
Tsung-Yi Lin, Michael Maire, Serge Belongie, James Hays, Pietro Perona, Deva
  Ramanan, Piotr Doll{\'a}r, and C~Lawrence Zitnick.
\newblock Microsoft coco: Common objects in context.
\newblock In {\em European conference on computer vision}, pages 740--755.
  Springer, 2014.

\bibitem{lin2011perceptual}
Weisi Lin and C-C~Jay Kuo.
\newblock Perceptual visual quality metrics: A survey.
\newblock {\em Journal of visual communication and image representation},
  22(4):297--312, 2011.

\bibitem{liu2011visual}
Hantao Liu and Ingrid Heynderickx.
\newblock Visual attention in objective image quality assessment: Based on
  eye-tracking data.
\newblock {\em IEEE Transactions on Circuits and Systems for Video Technology},
  21(7):971--982, 2011.

\bibitem{long2015fully}
Jonathan Long, Evan Shelhamer, and Trevor Darrell.
\newblock Fully convolutional networks for semantic segmentation.
\newblock In {\em Proceedings of the IEEE conference on computer vision and
  pattern recognition}, pages 3431--3440, 2015.

\bibitem{loshchilov2016sgdr}
Ilya Loshchilov and Frank Hutter.
\newblock Sgdr: Stochastic gradient descent with warm restarts.
\newblock {\em arXiv preprint arXiv:1608.03983}, 2016.

\bibitem{moorthy2009perceptually}
Anush~K Moorthy and Alan~C Bovik.
\newblock Perceptually significant spatial pooling techniques for image quality
  assessment.
\newblock In {\em Human Vision and Electronic Imaging XIV}, volume 7240, page
  724012. International Society for Optics and Photonics, 2009.

\bibitem{ninassi2006pseudo}
Alexandre Ninassi, Patrick Le~Callet, and Florent Autrusseau.
\newblock Pseudo no reference image quality metric using perceptual data
  hiding.
\newblock In {\em Human vision and electronic imaging XI}, volume 6057, page
  6057. International Society for Optics and Photonics, 2006.

\bibitem{ninassi2007does}
Alexandre Ninassi, Olivier Le~Meur, Patrick Le~Callet, and Dominique Barba.
\newblock Does where you gaze on an image affect your perception of quality?
  applying visual attention to image quality metric.
\newblock In {\em 2007 IEEE International Conference on Image Processing},
  volume~2, pages II--169. IEEE, 2007.

\bibitem{ostrovsky2005perceiving}
Yuri Ostrovsky, Patrick Cavanagh, and Pawan Sinha.
\newblock Perceiving illumination inconsistencies in scenes.
\newblock {\em Perception}, 34(11):1301--1314, 2005.

\bibitem{peirce2019psychopy2}
Jonathan Peirce, Jeremy~R Gray, Sol Simpson, Michael MacAskill, Richard
  H{\"o}chenberger, Hiroyuki Sogo, Erik Kastman, and Jonas~Kristoffer
  Lindel{\o}v.
\newblock Psychopy2: Experiments in behavior made easy.
\newblock {\em Behavior research methods}, 51(1):195--203, 2019.

\bibitem{peli1990contrast}
Eli Peli.
\newblock Contrast in complex images.
\newblock {\em JOSA A}, 7(10):2032--2040, 1990.

\bibitem{ponomarenko2013color}
Nikolay Ponomarenko, Oleg Ieremeiev, Vladimir Lukin, Karen Egiazarian, Lina
  Jin, Jaakko Astola, Benoit Vozel, Kacem Chehdi, Marco Carli, Federica
  Battisti, et~al.
\newblock Color image database tid2013: Peculiarities and preliminary results.
\newblock In {\em european workshop on visual information processing (EUVIP)},
  pages 106--111. IEEE, 2013.

\bibitem{radford2015unsupervised}
Alec Radford, Luke Metz, and Soumith Chintala.
\newblock Unsupervised representation learning with deep convolutional
  generative adversarial networks.
\newblock {\em arXiv preprint arXiv:1511.06434}, 2015.

\bibitem{robertson1977cie}
Alan~R Robertson.
\newblock The cie 1976 color-difference formulae.
\newblock {\em Color Research \& Application}, 2(1):7--11, 1977.

\bibitem{ronneberger2015u}
Olaf Ronneberger, Philipp Fischer, and Thomas Brox.
\newblock U-net: Convolutional networks for biomedical image segmentation.
\newblock In {\em International Conference on Medical image computing and
  computer-assisted intervention}, pages 234--241. Springer, 2015.

\bibitem{russell2008labelme}
Bryan~C Russell, Antonio Torralba, Kevin~P Murphy, and William~T Freeman.
\newblock Labelme: a database and web-based tool for image annotation.
\newblock {\em International journal of computer vision}, 77(1-3):157--173,
  2008.

\bibitem{scheirer2014perceptual}
Walter~J Scheirer, Samuel~E Anthony, Ken Nakayama, and David~D Cox.
\newblock Perceptual annotation: Measuring human vision to improve computer
  vision.
\newblock {\em IEEE transactions on pattern analysis and machine intelligence},
  36(8):1679--1686, 2014.

\bibitem{sheikh2006image}
Hamid~R Sheikh and Alan~C Bovik.
\newblock Image information and visual quality.
\newblock {\em IEEE Transactions on image processing}, 15(2):430--444, 2006.

\bibitem{sheikh2005information}
Hamid~R Sheikh, Alan~C Bovik, and Gustavo De~Veciana.
\newblock An information fidelity criterion for image quality assessment using
  natural scene statistics.
\newblock {\em IEEE Transactions on image processing}, 14(12):2117--2128, 2005.

\bibitem{sheikh2005live}
Hamid~R Sheikh, Zhou Wang, Lawrence Cormack, and Alan~C Bovik.
\newblock Live image quality assessment database release 2 (2005), 2005.

\bibitem{shi2015just}
Jianping Shi, Li Xu, and Jiaya Jia.
\newblock Just noticeable defocus blur detection and estimation.
\newblock In {\em Proceedings of the IEEE Conference on Computer Vision and
  Pattern Recognition}, pages 657--665, 2015.

\bibitem{shi2015visual}
Ran Shi, King~Ngi Ngan, Songnan Li, Raveendran Paramesran, and Hongliang Li.
\newblock Visual quality evaluation of image object segmentation: Subjective
  assessment and objective measure.
\newblock {\em IEEE Transactions on Image Processing}, 24(12):5033--5045, 2015.

\bibitem{talebi2018nima}
Hossein Talebi and Peyman Milanfar.
\newblock Nima: Neural image assessment.
\newblock {\em IEEE Transactions on Image Processing}, 27(8):3998--4011, 2018.

\bibitem{van1996perceptual}
Christian~J Van~den Branden~Lambrecht and Olivier Verscheure.
\newblock Perceptual quality measure using a spatiotemporal model of the human
  visual system.
\newblock In {\em Digital Video Compression: Algorithms and Technologies 1996},
  volume 2668, pages 450--461. International Society for Optics and Photonics,
  1996.

\bibitem{vu2008visual}
Cuong~T Vu, Eric~C Larson, and Damon~M Chandler.
\newblock Visual fixation patterns when judging image quality: Effects of
  distortion type, amount, and subject experience.
\newblock In {\em 2008 IEEE Southwest Symposium on Image Analysis and
  Interpretation}, pages 73--76. IEEE, 2008.

\bibitem{wallis2012image}
Thomas~SA Wallis and Peter~J Bex.
\newblock Image correlates of crowding in natural scenes.
\newblock {\em Journal of Vision}, 12(7):6--6, 2012.

\bibitem{wang2001designing}
Ching-Yang Wang, Shiuh-Ming Lee, and Long-Wen Chang.
\newblock Designing jpeg quantization tables based on human visual system.
\newblock {\em Signal Processing: Image Communication}, 16(5):501--506, 2001.

\bibitem{wang2007video}
Zhou Wang and Qiang Li.
\newblock Video quality assessment using a statistical model of human visual
  speed perception.
\newblock {\em JOSA A}, 24(12):B61--B69, 2007.

\bibitem{watson1983quest}
Andrew~B Watson and Denis~G Pelli.
\newblock Quest: A bayesian adaptive psychometric method.
\newblock {\em Perception \& psychophysics}, 33(2):113--120, 1983.

\bibitem{wichmann2001psychometric}
Felix~A Wichmann and N~Jeremy Hill.
\newblock The psychometric function: I. fitting, sampling, and goodness of fit.
\newblock {\em Perception \& psychophysics}, 63(8):1293--1313, 2001.

\bibitem{xiao2010sun}
Jianxiong Xiao, James Hays, Krista~A Ehinger, Aude Oliva, and Antonio Torralba.
\newblock Sun database: Large-scale scene recognition from abbey to zoo.
\newblock In {\em 2010 IEEE Computer Society Conference on Computer Vision and
  Pattern Recognition}, pages 3485--3492. IEEE, 2010.

\bibitem{xue2012understanding}
Su Xue, Aseem Agarwala, Julie Dorsey, and Holly Rushmeier.
\newblock Understanding and improving the realism of image composites.
\newblock {\em ACM Transactions on Graphics (TOG)}, 31(4):84, 2012.

\bibitem{yu2015just}
Aron Yu and Kristen Grauman.
\newblock Just noticeable differences in visual attributes.
\newblock In {\em Proceedings of the IEEE International Conference on Computer
  Vision}, pages 2416--2424, 2015.

\bibitem{yu2015multi}
Fisher Yu and Vladlen Koltun.
\newblock Multi-scale context aggregation by dilated convolutions.
\newblock {\em arXiv preprint arXiv:1511.07122}, 2015.

\bibitem{zhang2019aet}
Liheng Zhang, Guo-Jun Qi, Liqiang Wang, and Jiebo Luo.
\newblock Aet vs. aed: Unsupervised representation learning by auto-encoding
  transformations rather than data.
\newblock In {\em Proceedings of the IEEE Conference on Computer Vision and
  Pattern Recognition}, pages 2547--2555, 2019.

\bibitem{zhang2018unreasonable}
Richard Zhang, Phillip Isola, Alexei~A Efros, Eli Shechtman, and Oliver Wang.
\newblock The unreasonable effectiveness of deep features as a perceptual
  metric.
\newblock In {\em Proceedings of the IEEE Conference on Computer Vision and
  Pattern Recognition}, pages 586--595, 2018.

\end{thebibliography}
}

\end{document}